\documentclass{article}
\usepackage{spconf,amsmath,graphicx,amssymb,booktabs,multirow}

\usepackage{graphicx}

\title{\textbf{CM-PIE: CROSS-MODAL PERCEPTION FOR INTERACTIVE-ENHANCED \\AUDIO-VISUAL VIDEO PARSING}}
\name{Yaru Chen$^{1,4}$,
      \text{Ruohao Guo}$^{2}$, %\sthanks{Thanks to XYZ agency for funding.}$,
      \text{Xubo Liu}$^{1}$, 
      Peipei Wu$^{1}$,
      Guangyao Li$^{3}$,
      Zhenbo Li$^{4}$, 
      Wenwu Wang$^{1}$
      }
\address{$^{1}$Centre for Vision Speech and Signal Processing~(CVSSP), University of Surrey, United Kindom\\
	$^{2}$School of Intelligence Science and Technology, Peking University, China\\
	$^{3}$Gaoling School of Artificial Intelligence, Renmin University of China, China\\
    $^{4}$College of Information and Electrical Engineering, China Agricultural University, China
}

\begin{document}
%\ninept
%
\maketitle
\begin{abstract}
Audio-visual video parsing is the task of categorizing a video at the segment level with weak labels, and predicting them as audible or visible events. Recent methods for this task leverage the attention mechanism to capture the semantic correlations among the whole video across the audio-visual modalities. However, these approaches have overlooked the importance of individual segments within a video and the relationship among them, and tend to rely on a single modality when learning features. In this paper, we propose a novel interactive-enhanced cross-modal perception method~(CM-PIE), which can learn fine-grained features by applying a segment-based attention module. Furthermore, a cross-modal aggregation block is introduced to jointly optimize the semantic representation of audio and visual signals by enhancing inter-modal interactions. The experimental results show that our model offers improved parsing performance on the Look, Listen, and Parse dataset compared to other methods.
\end{abstract}
\section{Introduction}
\label{sec:intro}

Humans perceive multisensory signals through sight, hearing, touch and more, acquiring multi-modal information when they explore the environment.~Enabling machines to fuse multi-modal information like humans to fuse better scene perception and understanding is a valuable research topic~\cite{malik2021automatic,wei2022learning}. As two basic modalities, audio and visual play a vital role in machine perception and understanding of scenes~\cite{gao2019co, li2022learning}. Some researchers used audio and visual signals to capture the comprehensive scene information, which can improve model performance and generalization~\cite{tian2018audio,wu2019dual}. However, the above methods usually assume audio and visual signals are temporally aligned, which, however, may not be the case, and thereby leading to inaccuracies in parsing the video.
\begin{figure}[htb]
\begin{minipage}[b]{1.0\linewidth}
  \centering
  \centerline{\includegraphics[width=8.7cm]{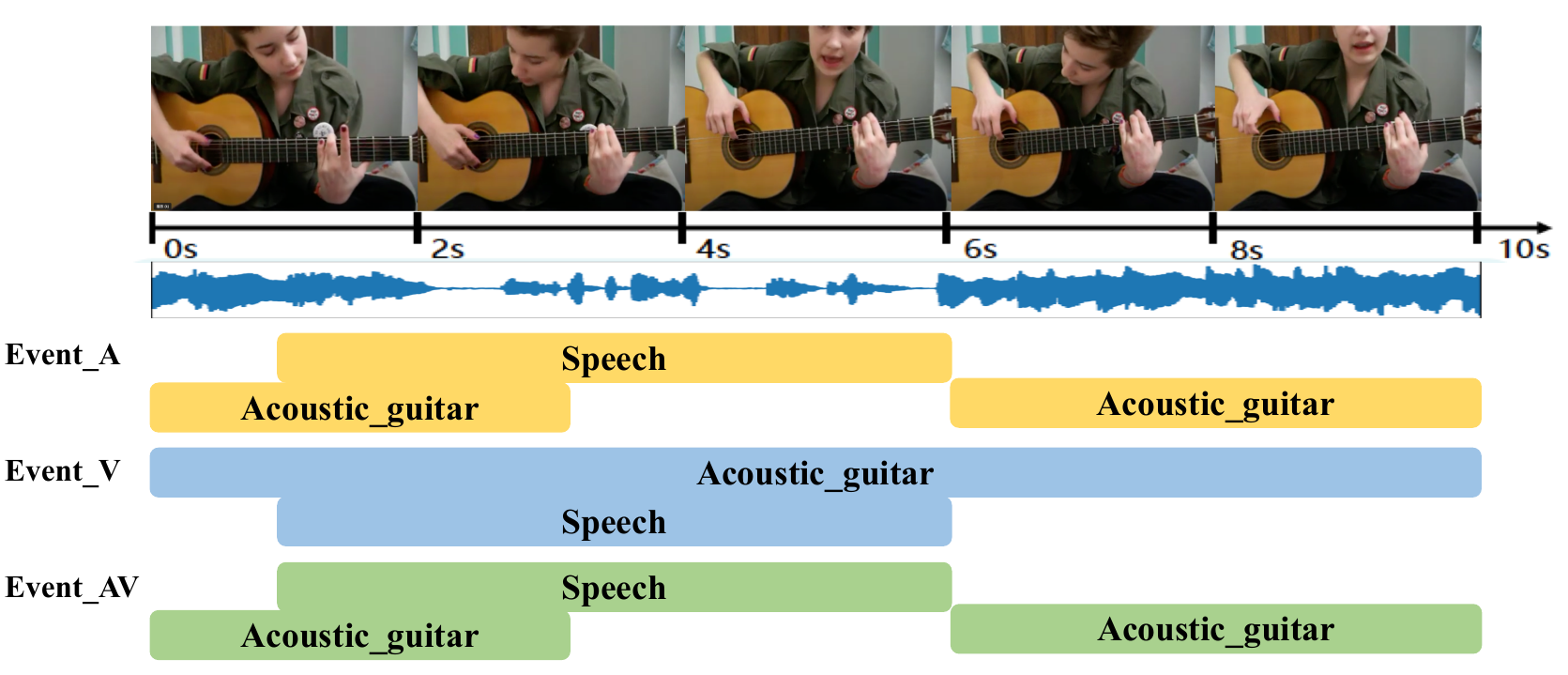}}
%  \vspace{2.0cm}
  % \centerline{(a) Result 1}\medskip
\end{minipage}
\caption{Example of the AVVP task. Taking the audio and visual data as input, the task is to determine the event categories, their temporal boundaries and the modality of the event. Note that it is possible for audio events and visual events to be asynchronous~(e.g.~acoustic guitar).}
\label{fig:res}
\end{figure} 

To solve this problem, Tian et al~\cite{tian2020unified} proposed the audio-visual video parsing (AVVP) task for a more fine-grained scene understanding. As shown in Fig.~1, AVVP aims to locate the temporal boundaries of event categories within a video with weak labels, and annotate them as audible, visible, or a combination of both. This task involves two challenges: One is to predict the event by extracting useful information from every segment. The other is to aggregate the cross-modality information to parse audio and visual events based on weak labels.

In~\cite{tian2020unified}, a method combining hybrid attention networks (HAN) and a multi-modal multiple instance learning (MMIL) is used to aggregate multi-modal temporal contexts, together with the identification and suppression of noisy labels for each modality. Subsequently, Yu et al~\cite{yu2022mm} proposed a method to capture and integrate multimodal pyramid features in different temporal scales. Afterward, Chen et al~\cite{chen2023cm} explored common and specific characters between 2D and 3D visual features, and visual and audio features separately. While the above methods have achieved promising improvements, there are still some limitations: 1) The existing approaches explore all features holistically from a whole video, but overlook the importance of individual segments in a video and the relationship among them. 2) The previous methods may cause modality bias~\cite{odegaard2015biases} due to their ineffective fusion of information from different modalities.

To address the above issues, we propose a novel interactive-enhanced cross-modal perception method that leverages the advantage of audio and visual modality. In detail, two stages are involved. Firstly, we propose a segment-based attention~(SA) module, which can effectively learn the importance of each segment and capture the relationship between different segments in the whole video. Secondly, we design a cross-modal aggregation~(CMA) block, which can enrich feature representation and enhance the ability of the model to parse video. Experimental results on the benchmark dataset show that our method achieves significant improvements as compared with the baseline method, where the event-level audio-visual event metric is improved from 48.0\% to 51.3\%.

The remainder of this paper is organized as follows. The next section introduces the SA module and CMA block we proposed for efficient video parsing. Section 3 presents the experimental settings and the evaluation results. Conclusion and future directions are given in Section 4.  
% While these methods have indeed shown promising improvements, there are two problems that need to be resolved. One is the above methods always explore all features in a video from a global perspective and don't analyze the importance of local features to the video. Another one is previous methods used a single cross-attention to fuse features, which would lead the model to lack correlation information between modalities and easily cause modality bias.

\begin{figure*}
    \begin{minipage}[b]{1.0\linewidth}
  \centering
  \centerline{\includegraphics[width=16.5cm]{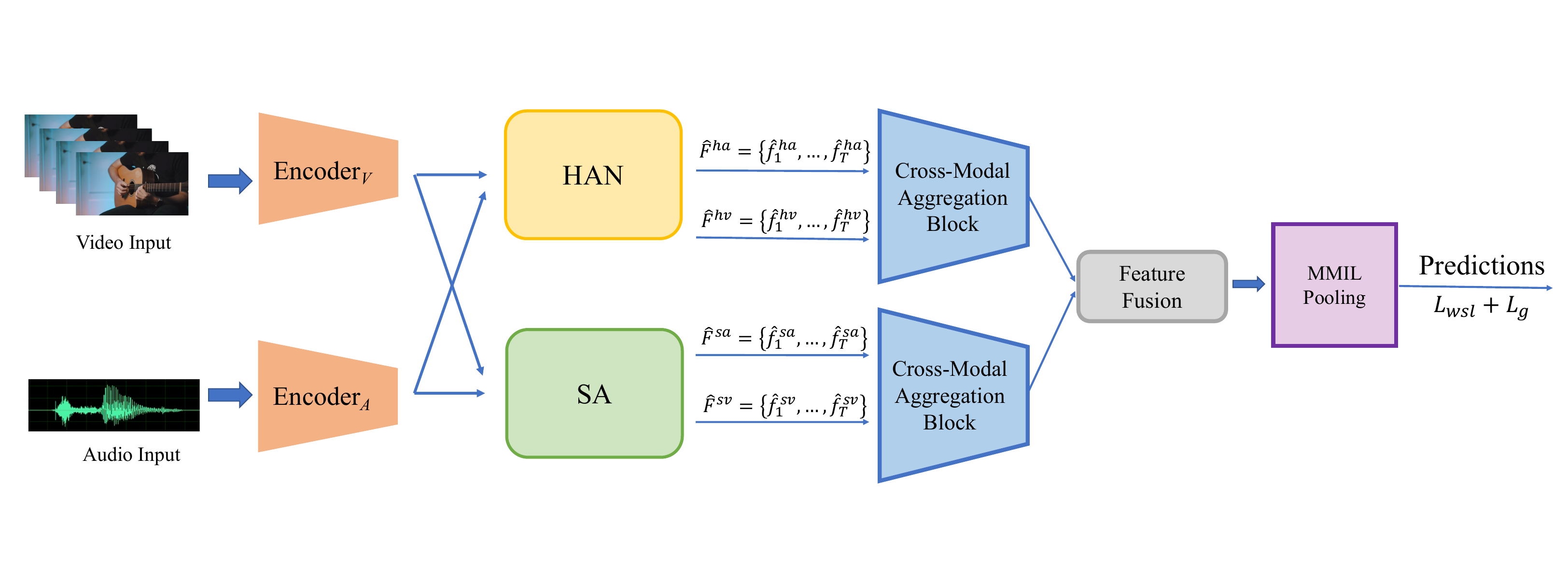}}
 % \vspace{2.0cm}
 %  \centerline{(a) Result 1}\medskip
\end{minipage}
\vspace{-1.2cm}
\caption{The pipeline of our proposed interactive-enhanced cross-modal perception model (CM-PIE). It uses pre-trained encoders to extract audio and visual features. Firstly, we used two attention-based modules to learn fine-grained information. Then two cross-modal aggregation blocks are used to improve feature representation. Finally, multi-modal fusion is exploited and using MMIL Pooling to get the video-level event prediction.}
\label{fig:res}
\end{figure*}

\section{PROPOSED METHOD}
\label{sec:format}

\subsection{Problem Statement}
The AVVP task aims to identify the event of every segment into audio event, visual event and audio-visual event, together with their classes. When we input an audio-visual video sequence with \(T\) seconds, we regard the video sequence as divided into \(T\) segments with each segment lasting for one second long, expressed as \(S=\{A_{t},V_{t}\}_{t=1}^{T}\), where \(A\) and \(V\) denote the audio and visual segment pairs in time \(t\). We use \(y_t^a\in\mathbb{R}^{ C}\), \(y_t^v\in\mathbb{R}^{C}\) and \(y_t^{av}\in\mathbb{R}^{C}\) to represent the audio, visual and audio-visual event labels at time \(t\), where \(C\) is the number of event categories. The audio-visual event occurs when the audio event and visual event happen at the same time, which means \(y_t^{at} = y_t^a * y_t^v\). Noted that we only have weak labels for training, but have detailed event labels with modalities and temporal boundaries for evaluation.

\subsection{Segment-based Attention}

As shown in Fig.~2, we obtain the audio and visual features by using pre-trained audio and visual encoders, denoted as \( \{f^a_t\} ^T_{t=1}\), \( \{f^v_t\} ^T_{t=1}\). These features are firstly aggregated by the hybrid attention network (HAN)~\cite{tian2020unified}, which utilizes self-attention and cross-modal attention to obtain the intra- and cross-modality information. These multi-head attention blocks \(\delta_{attn}\) can be described as follows:
\begin{equation}
\small
    \delta_{attn}(Q,K,V) = Softmax(\frac{QK^T}{\sqrt{d}})V\
\end{equation}
where \(Q\), \(K\), \(V\) are query, key and value, with \(d\) being the dimension of the vector \(Q\). The process of obtaining features from HAN can be denoted as:
\begin{equation}
    \small
    \hat{f}^{ha}_t = f^{ha}_t + \delta_{attn}(f^{ha}_t, F^{ha}, F^{ha}) + \delta_{attn}(f^{ha}_t, F^{hv}, F^{hv})
\end{equation}
\begin{equation}
    \small
    \hat{f}^{hv}_t = f^{hv}_t + \delta_{attn}(f^{hv}_t, F^{hv}, F^{hv}) + \delta_{attn}(f^{hv}_t, F^{ha}, F^{ha})
\end{equation}
where \(f^{ha}_t\) and \(f^{hv}_t\) are the feature vectors at a specific time \(t\) extracted from audio and visual encoders. \(F^{ha}\) and \(F^{hv}\) stand for the feature set in the same video, which are defined as \(F^{ha}=\{f^{ha}_1, ..., f^{ha}_T\}\in\mathbb{R}^{T\times{d}}\) and \(F^{hv}=\{f^{hv}_1, ..., f^{hv}_T\}\in\mathbb{R}^{T\times{d}}\). \(\hat{f}^{ha}_t\) and \(\hat{f}^{hv}_t\) are aggregated features obtained from HAN, and \(d\) is the feature dimension which is set to 512 in this paper.

The HAN module considers holistic feature information but ignores the features from different segments~\cite{tian2020unified}.~To address this limitation, we propose the segment-based attention~(SA) module to obtain fine-grained feature information, which is similar to channel attention mechanism~\cite{hu2018squeeze}, which can not only selectively enhance or weaken different segments to highlight important feature information, but also assist the model in capturing feature relationships among different segments. 

As shown in Fig.~3, in the SA module, firstly, the average feature representation is computed along the segment dimension, then, these representations are input into a well-defined neural network to generate the segment-based attention weight matrix \(W^a_t\) and \(W^v_t\):
\begin{equation}
\small
    W^a_t = \varphi(\sum_{s=1}^T(\phi^a_{t}))
\end{equation}
\begin{equation}
\small
    W^v_t = \varphi(\sum_{s=1}^T(\phi^v_{t}))
\end{equation}
where \(\phi^a_{t}\) and \(\phi^v_{t}\) denote audio and visual features within a video, whose dimensions are (\(b\),~\(s\),~\(d\)), representing batch size, segments, and dimension, respectively, and \(\varphi\) indicates a neural network that includes several linear layers and activation functions~(e.g. \(ReLU\) and \(Sigmoid\)). These attention weights enable us to modulate the feature representations of each segment according to their local importance. Afterward, we multiply the input features with the attention weights to obtain refined features \(\Tilde{f}^a_t\) and \(\Tilde{f}^a_t\) in terms of their importance, 
\begin{equation}
    \Tilde{f}^a_t = \phi^a_{t} * W^a_t
\end{equation}
\begin{equation}
    \Tilde{f}^v_t = \phi^v_{t} * W^v_t
\end{equation}
Subsequently, these features are input into the HAN module to do further aggregation to get \(\hat{f}^{sa}_t\) and \(\hat{f}^{sv}_t\).

\subsection{Cross-modal Aggregation}
Two cross-modal aggregation~(CMA) blocks are used to facilitate the learning of the correlations between audio and visual features. As shown in formula~(2) and (3), previous method~\cite{tian2020unified,chen2023cm} only used single modality as the input of vector \(K\) and \(V\), which leads to sub-optimal cross-modal fusion results. In contrast, we concatenate audio and visual features, and then, the single-modality features are brought closer to the fused features. This can mitigate the impact of modality bias and enhance the effectiveness of cross-modal information fusion:
\begin{equation}
    \hat{g}^{ha}_t = \delta_{attn}(\hat{f}^{ha}_t, \hat{F}^{ha}\oplus\hat{F}^{hv}, \hat{F}^{ha}\oplus\hat{F}^{hv})
\end{equation}
\begin{equation}
    \hat{g}^{hv}_t = \delta_{attn}(\hat{f}^{hv}_t, \hat{F}^{ha}\oplus\hat{F}^{hv}, \hat{F}^{ha}\oplus\hat{F}^{hv})
\end{equation}
\begin{equation}
    \hat{g}^{sa}_t = \delta_{attn}(\hat{f}^{sa}_t, \hat{F}^{sa}\oplus\hat{F}^{sv}, \hat{F}^{sa}\oplus\hat{F}^{sv})
\end{equation}
\begin{equation}
    \hat{g}^{sv}_t = \delta_{attn}(\hat{f}^{sv}_t, \hat{F}^{sa}\oplus\hat{F}^{sv}, \hat{F}^{sa}\oplus\hat{F}^{sv})
\end{equation}
where \(\hat{F}^{ha}\) and \(\hat{F}^{hv}\) are the sets of aggregated audio and visual feature obtained from HAN as defined in the previous section, \(\hat{F}^{sa}\) and \(\hat{F}^{sv}\) are the sets of the audio and visual features obtained from SA, \(\hat{g}^{ha}_t\) and \(\hat{g}^{hv}_t\) are the features obtained after applying the CMA block, and \(\oplus\) means concatenating operation. In the same way, \(\hat{g}^{sa}_t\), \(\hat{g}^{sv}_t\) are the aggregation features derived from the SA module. Subsequently, we perform feature fusion operations as follows:
\begin{equation}
    \Tilde{g}^a_t = Mean(~\hat{g}^{ha}_t,~\hat{g}^{sa}_t~)
\end{equation}
\begin{equation}
    \Tilde{g}^v_t = Mean(~\hat{g}^{hv}_t,~\hat{g}^{sv}_t~)
\end{equation}
where \textit{Mean} denotes taking the average of the two vectors element-wise. With the feature \(\Tilde{g}^a_t\) and \(\Tilde{g}^a_t\), we can obtain the segment-wise event prediction, which can be turned into video-level predictions by a pooling method, such as MMIL Pooling~\cite{tian2020unified} based on a shared fully-connected layer and an activation function.

\section{EXPERIMENTAL RESULTS}

\begin{table*}[ht]
\begin{center}

\caption{Comparison with the state-of-the-art methods on the LLP dataset in terms of F-scores. The event-level F-scores are calculated using a threshold of mIoU = 0.5.}
\vspace{0.5em}
\scalebox{0.99}{

\begin{tabular}{c|ccccc|ccccc}
\hline
\multirow{2}{*}{Method} & \multicolumn{5}{c|}{Segment-level}  & \multicolumn{5}{c}{Event-level}     \\ 
\cline{2-11} 
                        & Audio & Visual & AV & Ty@AV & Ev@AV & Audio & Visual & AV & Ty@AV & Ev@AV \\ 
                        \hline
AVE~\cite{tian2018audio}     & 47.2  & 37.1   & 35.4  & 39.9  & 41.6  & 40.4 & 34.7 & 31.6 & 35.5 & 36.5  \\
AVSDN~\cite{lin2019dual}     & 47.8  & 52.0   & 37.1  & 45.7  & 50.8  & 34.1 & 46.3 & 26.5 & 35.6 & 37.7  \\
HAN~\cite{tian2020unified}   & 60.1  & 52.9   & 48.9  & 54.0  & 55.4  & 51.3 & 48.9 & 43.0 & 47.7 & 48.0  \\
MM-Pyramid~\cite{yu2022mm}   & 60.9  & 54.4   & 51.8  & 55.1  & \textbf{57.6}  & 52.7 & 50.0 & 44.4 & 49.9 & 50.5  \\
CM-CS+HAN~\cite{chen2023cm}  & 57.1  & \textbf{57.6}   & \textbf{52.9}  & \textbf{55.9}  & 53.9  & 49.4 & \textbf{54.2} & \textbf{46.7} & \textbf{50.1} & 47.8  \\
\hline
\textbf{CM-PIE (Ours) }      & \textbf{61.7}  & 55.2  & 50.1  & 55.7  & 56.8 & \textbf{53.7} & 51.3 & 43.6 & 49.5 & \textbf{51.3}   \\ 
\hline
\end{tabular}

}

\vspace{-1.5em}
\label{cmp}
\end{center}
\end{table*}

\subsection{Experiment Setup}
\begin{figure}[ht]
\begin{minipage}[b]{1.0\linewidth}
  \centering
  \centerline{\includegraphics[width=6.8cm]{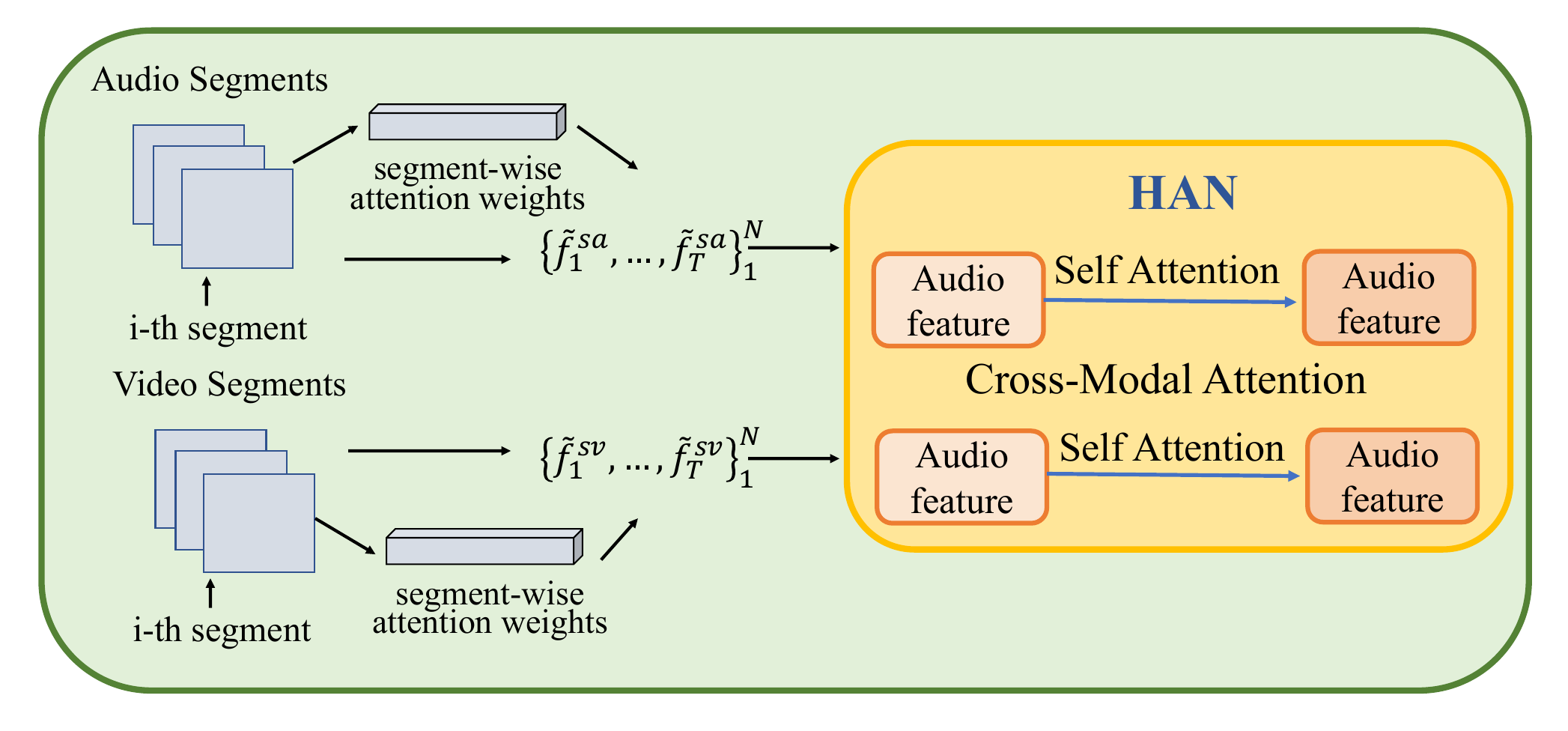}}
%  \vspace{2.0cm}
  % \centerline{(a) Result 1}\medskip
\end{minipage}
\caption{Details of segment-based attention~(SA) module. This module can learn the importance of each segment by calculating the segment-wise attention weights.}
\label{fig:res}
\end{figure}

\textbf{Dataset.} The LLP dataset~\cite{tian2020unified} is used to evaluate our method. This dataset has 11849 videos with 25 categories taken from YouTube, containing various scenes and species. The dataset has 10000 videos with weak labels as the training set, 1200 videos and 649 videos as the testing set and the validation set with fully annotated labels. Each video has 10 segments and each segment lasts 1 second.\\
\textbf{Implementation details.} We use pre-trained VGGish~\cite{hershey2017cnn} to get 128-D audio features, and use 2D and 3D ResNet~\cite{he2016deep, tran2018closer} to extract 512-D visual features. The concatenation of 2D and 3D visual features is subsequently fed through a multi-layer perceptron (MLP) to generate segment-wise representations.\\
\textbf{Evaluation Metrics.} Following~\cite{tian2020unified}, we evaluate the performance of the proposed methods using F-scores, calculated at both segment-level and event-level. For the segment-level performance, we compute the F-score for each segment. The F-scores for the Audio, Visual and AV columns in Table 1 were calculated by comparing the predictions with the ground truth annotations for audio, visual and audio-visual sequence (as shown in Fig. 1), and averaged for all the video sequences. With Ty@AV, we evaluate the overall performance by averaging the results from Audio, Visual, and AV columns in Table 1. With Ev@AV, we evaluate the performance of the models for event classification by averaging of the F-scores calculated for each event, i.e. comparing the prediction results for each event and its ground truth annotation along the segments in the video sequences. For the event-level performance, we first concatenate the positive segments in a sequential order, then calculate the F-score for each event, finally take the average F-score for all the events. The event-level evaluations also take into account the accuracy of the onset and offset time of event occurrence.  
\vspace{-0.4cm}
\subsection{Comparison with State-of-the-art Method}
We compare our method with several popular baselines, such as HAN~\cite{tian2020unified}, MM-Pyramid~\cite{yu2022mm} and CM-CS~\cite{chen2023cm}. We also compare our method to some modified audio-visual event localization methods, including AVE~\cite{tian2018audio} and AVSDN~\cite{lin2019dual}. Table 1 shows the results. The proposed method shows improved performance, due to the use of interactive fusion of the audio and visual information. To further demonstrate this, we present results for adding the SA module and CMA block. The outcomes demonstrate a significant performance improvement of our model over the baseline method~(HAN) across all evaluation metrics. Compared with HAN, our model is improved in both single and multi-modal metrics. For example, it achieves up to a 1.6\% improvement in Audio~\&~Segment-level~(61.7\%~vs.~60.1\%) and 2.3\% enhancement in Visual~\&~Segment-level~(55.2\%~vs.~52.9\%). What's more, our model achieves a 3.3\% improvement in the Ev@AV~\&~Event-level metric~(51.3\%~vs.~48.0\%). This confirms that accurate event localization at different segments can be achieved by learning useful information from important segments and the fusion of features of different modalities, thereby improving video parsing performance. Compared with other methods, our method achieves state-of-the-art results on some indicators, such as Audio~\&~Segment-level and Ev@AV~\&~Event-level, and the model also shows promising results on other performance indicators.
\begin{table}[h]
\centering
\caption{Ablation Study on the LLP dataset. Seg Attention denotes adding the SA module. \textit{w/o.~V} and \textit{w/o.~A} means using only the CMA block for audio features or visual features.}
\vspace{0.5em}
\label{tab:noiseless}
\small
\begin{tabular}{@{}cccccc@{}}
\toprule

\multicolumn{6}{c}{Segment-level} \\ \midrule
Methods & Audio & Visual & AV & Ty@AV & Ev@AV \\ \midrule
HAN~\cite{tian2020unified} & {60.1} & {52.9} & 48.9 & 54.0 & 55.4 \\ 
+Seg Attention & {60.7} & \textbf{{55.5}} & 48.6 & 54.9 & 56.1 \\
\textit{~w/o.~V} & {60.4} & {55.3} & \textbf{51.2} & 55.6 & 56.1 \\
\textit{~w/o.~A} & {61.5} & {54.8} & 50.0 & 55.4 & \textbf{57.0} \\
\textbf{CM-PIE~(Ours)} & \textbf{{61.7}} & {55.2} & 50.1 & \textbf{55.7} & 56.8 \\  \midrule
\multicolumn{6}{c}{Event-level} \\  \midrule
HAN~\cite{tian2020unified} & 51.3 & 48.9 & {43.0} & {47.7} & 48.0 \\
+Seg Attention & {53.2} & {49.8} & 42.1 & 48.3 & 50.5 \\
\textit{~w/o.~V} & {52.2} & {50.0} & \textbf{43.8} & 48.6 & 49.3 \\ 
\textit{~w/o.~A} & {53.0} & {51.1} & 43.5 & 49.2 & 50.9 \\
\textbf{CM-PIE~(Ours)} & \textbf{{53.7}} & \textbf{{51.3}} & 43.6 & \textbf{49.5} &\textbf{{ 51.3}} \\
\bottomrule
\end{tabular}
\end{table}
\vspace{-0.4cm}
\subsection{Ablation Study}
Ablation studies aim to examine the influence of each part within the proposed approach and the results are shown in Table 2. We notice that both the SA module and the CMA block can improve the experimental results in several metrics. By learning from crucial segments, we have achieved notable performance enhancements in both visual and audio-visual evaluations.~This module can address the potential limitation described earlier of relying solely on aggregated features from the HAN. Furthermore, the usage of the CMA block significantly improves the results in terms of Ty@AV and Ev@AV evaluations. Given that the Ev@AV evaluation considers the F-score for all audio and visual events, the enhancement in Ev@AV indicates the substantial improvement given by the proposed method in video parsing.
%\vspace{-0.1cm}
\section{CONCLUSION}
In this paper, we have presented a novel weakly-supervised audio-visual video parsing framework. Two modules are introduced to leverage the segment relationships and semantics across the modalities. The segment-based attention module extracts local features from segments, which can learn more fine-grained semantics in a video. The cross-modal aggregation block effectively reduces modality bias, facilitating the effective fusion of the cross-modal information. Our approach has achieved promising results on the LLP dataset. In future work, we will further study the relationship between different segments across the video sequence.
\label{sec:typestyle}

\vfill\pagebreak
\label{sec:refs}

\bibliographystyle{IEEEtran}
\bibliography{strings,refs}

\begin{thebibliography}{10}
\providecommand{\url}[1]{#1}
\csname url@samestyle\endcsname
\providecommand{\newblock}{\relax}
\providecommand{\bibinfo}[2]{#2}
\providecommand{\BIBentrySTDinterwordspacing}{\spaceskip=0pt\relax}
\providecommand{\BIBentryALTinterwordstretchfactor}{4}
\providecommand{\BIBentryALTinterwordspacing}{\spaceskip=\fontdimen2\font plus
\BIBentryALTinterwordstretchfactor\fontdimen3\font minus
  \fontdimen4\font\relax}
\providecommand{\BIBforeignlanguage}[2]{{%
\expandafter\ifx\csname l@#1\endcsname\relax
\typeout{** WARNING: IEEEtran.bst: No hyphenation pattern has been}%
\typeout{** loaded for the language `#1'. Using the pattern for}%
\typeout{** the default language instead.}%
\else
\language=\csname l@#1\endcsname
\fi
#2}}
\providecommand{\BIBdecl}{\relax}
\BIBdecl

\bibitem{malik2021automatic}
M.~Malik, M.~K. Malik, K.~Mehmood, and I.~Makhdoom, ``{Automatic speech
  recognition: a survey},'' \emph{Multimedia Tools and Applications}, vol.~80,
  pp. 9411--9457, 2021.

\bibitem{wei2022learning}
Y.~Wei, D.~Hu, Y.~Tian, and X.~Li, ``{Learning in audio-visual context: A
  review, analysis, and new perspective},'' \emph{arXiv preprint
  arXiv:2208.09579}, 2022.

\bibitem{gao2019co}
R.~Gao and K.~Grauman, ``{Co-separating sounds of visual objects},'' in
  \emph{Proceedings of the IEEE/CVF International Conference on Computer
  Vision}, 2019, pp. 3879--3888.

\bibitem{li2022learning}
G.~Li, Y.~Wei, Y.~Tian, C.~Xu, J.-R. Wen, and D.~Hu, ``{Learning to answer
  questions in dynamic audio-visual scenarios},'' in \emph{Proceedings of the
  IEEE/CVF Conference on Computer Vision and Pattern Recognition}, 2022, pp.
  19\,108--19\,118.

\bibitem{tian2018audio}
Y.~Tian, J.~Shi, B.~Li, Z.~Duan, and C.~Xu, ``{Audio-visual event localization
  in unconstrained videos},'' in \emph{Proceedings of the European Conference
  on Computer Vision (ECCV)}, 2018, pp. 247--263.

\bibitem{wu2019dual}
Y.~Wu, L.~Zhu, Y.~Yan, and Y.~Yang, ``{Dual attention matching for audio-visual
  event localization},'' in \emph{Proceedings of the IEEE/CVF International
  Conference on Computer Vision}, 2019, pp. 6292--6300.

\bibitem{tian2020unified}
Y.~Tian, D.~Li, and C.~Xu, ``{Unified multisensory perception:
  Weakly-supervised audio-visual video parsing},'' in \emph{Proceedings of the
  16th European Conference on Computer Vision, Glasgow, UK, August 23--28,
  2020}.\hskip 1em plus 0.5em minus 0.4em\relax Springer, 2020, pp. 436--454.

\bibitem{yu2022mm}
J.~Yu, Y.~Cheng, R.-W. Zhao, R.~Feng, and Y.~Zhang, ``{MM-Pyramid: Multimodal
  pyramid attentional network for audio-visual event localization and video
  parsing},'' in \emph{Proceedings of the 30th ACM International Conference on
  Multimedia}, 2022, pp. 6241--6249.

\bibitem{chen2023cm}
H.~Chen, D.~Zhu, G.~Zhang, W.~Shi, X.~Zhang, and J.~Li, ``{CM-CS: cross-modal
  common-specific feature learning for audio-visual video parsing},'' in
  \emph{IEEE International Conference on Acoustics, Speech and Signal
  Processing (ICASSP)}.\hskip 1em plus 0.5em minus 0.4em\relax IEEE, 2023, pp.
  1--5.

\bibitem{odegaard2015biases}
B.~Odegaard, D.~R. Wozny, and L.~Shams, ``{Biases in visual, auditory, and
  audiovisual perception of space},'' \emph{PLoS Computational Biology},
  vol.~11, no.~12, p. e1004649, 2015.

\bibitem{hu2018squeeze}
J.~Hu, L.~Shen, and G.~Sun, ``{Squeeze-and-excitation networks},'' in
  \emph{Proceedings of the IEEE Conference on Computer Vision and Pattern
  Recognition}, 2018, pp. 7132--7141.

\bibitem{lin2019dual}
Y.~Lin, Y.~Li, and Y.~F. Wang, ``{Dual-modality seq2seq network for
  audio-visual event localization},'' in \emph{IEEE International Conference on
  Acoustics, Speech and Signal Processing (ICASSP)}.\hskip 1em plus 0.5em minus
  0.4em\relax IEEE, 2019, pp. 2002--2006.

\bibitem{hershey2017cnn}
S.~Hershey, S.~Chaudhuri, D.~P. Ellis, J.~F. Gemmeke, A.~Jansen, R.~C. Moore,
  M.~Plakal, D.~Platt, R.~A. Saurous, B.~Seybold \emph{et~al.}, ``{CNN
  architectures for large-scale audio classification},'' in \emph{2017 IEEE
  International Conference on Acoustics, Speech and Signal Processing
  (ICASSP)}.\hskip 1em plus 0.5em minus 0.4em\relax IEEE, 2017, pp. 131--135.

\bibitem{he2016deep}
K.~He, X.~Zhang, S.~Ren, and J.~Sun, ``{Deep residual learning for image
  recognition},'' in \emph{Proceedings of the IEEE Conference on Computer
  Vision and Pattern Recognition}, 2016, pp. 770--778.

\bibitem{tran2018closer}
D.~Tran, H.~Wang, L.~Torresani, J.~Ray, Y.~LeCun, and M.~Paluri, ``{A closer
  look at spatiotemporal convolutions for action recognition},'' in
  \emph{Proceedings of the IEEE Conference on Computer Vision and Pattern
  Recognition}, 2018, pp. 6450--6459.

\end{thebibliography}

\end{document}